\documentclass[10pt,twocolumn,letterpaper]{article}

\usepackage{times}
\usepackage{epsfig}
\usepackage{graphicx}
\usepackage{amsmath}
\usepackage{amssymb}

\usepackage[breaklinks=true,bookmarks=false]{hyperref}

\begin{document}

\title{Convex Decomposition of Indoor Scenes}

\author{Vaibhav Vavilala\\
UIUC\\
\and
David Forsyth\\
UIUC\\
}

\maketitle

\begin{abstract}
We describe a method to parse a complex, cluttered indoor scene into primitives which offer a parsimonious abstraction of scene structure.  Our primitives are simple convexes. Our method uses a learned regression procedure to parse a scene into a fixed number of convexes from RGBD input, and can optionally accept segmentations to improve the decomposition.  The result is then polished with a descent method which adjusts the convexes to produce a very good fit, and greedily removes superfluous primitives. Because the entire scene is parsed, we can evaluate using traditional depth, normal, and segmentation error metrics. Our evaluation procedure demonstrates that the error from our primitive representation is comparable to that of predicting depth from a single image.
\end{abstract}

\section{Introduction}
\label{sec:intro}

Primitive based abstractions of geometry offer simpler reasoning -- it is easier, for example, to deal with a fridge as a cuboid than attend to the detailed curves around a corner. But obtaining primitive representations that abstract usefully and accurately has been hard.  Parsing isolated objects into parts has seen considerable recent progress, but parsing indoor scenes into collections of primitives has remained difficult. We describe a method to decompose indoor scenes into convex primitives which largely describe the scene. Our method is trained on labeled RGBD + semantic segmentation maps, and the predicted decomposition can be refined on depth + segmentation maps that are predicted from a suitable pretrained network. 

Our method draws on a long history of fitting primitives to objects with two main methods (review Sec.~\ref{sec:related}). A {\bf descent method} chooses primitives for a given geometry by minimizing a cost function. These methods are plagued by local minima and robustness issues. A {\bf regression method} uses a learned predictor to map geometry to primitives
and their parameters. These methods can pool examples to avoid local minima, but may not get the best prediction for a given input. Our method uses regression to predict an initial decomposition, then applies descent to polish the primitives for each particular test geometry. We show that both steps are critical to achieving a good fit.   

\begin{figure}[t!]
\centering
\begin{tabular}{@{}c@{}}
  \includegraphics[width=0.95\linewidth]{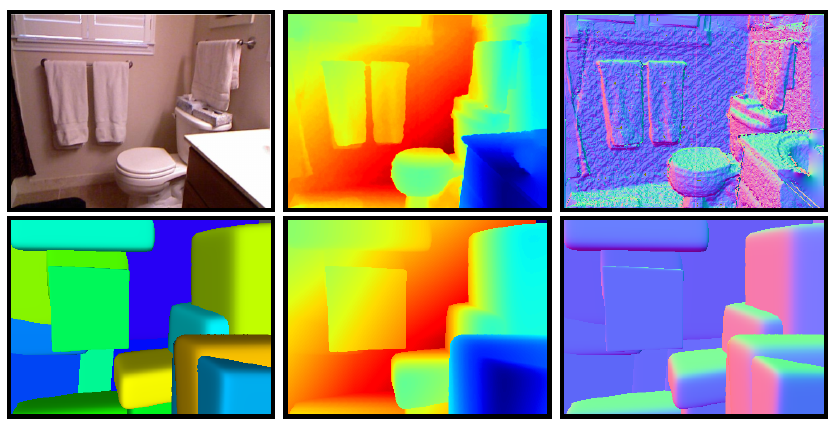} 
  \label{fig:teas1}
  \end{tabular}
  \caption{Our method decomposes complex cluttered indoor scenes into simple geometric shapes. The segmentation itself is supervised from known labels, and the losses attempt to capture the geometric information in the depth map input with a pre-defined number of convexes. A convex decomposition predicted by a neural network is the input to a refinement process that removes excess parts while improving the fit. \textit{Top row}: Test RGB image, depth map input, ground truth normals. \textit{Bottom row}: predicted segmentation, predicted depth map, predicted normals. Notice how strongly the predicted depth and normals match the ground truth. Different from previous work, our method generally captures the whole input, allowing traditional depth and normal error metrics.}
\vspace{-4mm}
\end{figure}

It can be hard to tell if a primitive based representation captures the abstraction one wants.
But for a representation to be useful, it must have three necessary properties:

\begin{enumerate}
\item \textbf{Accuracy}: The primitives should represent the original geometry.  As Section~\ref{sec:exper} demonstrates, we do not need to create metrics specifically for scene parsing and instead existing evaluation procedures from depth/segmentation prediction literature are sufficient.
\item \textbf{Segmentation}:  Generally, the decomposition should ``make sense.''  For isolated objects, this means that each primitive should correspond to an object part (which is difficult to evaluate). For scenes, this means that each primitive should span no more than one object.  An object might be explained by multiple primitives, but a primitive should not explain multiple objects. This property is not usually evaluated quantitatively; we show that it can be.
\item \textbf{Parsimony}: In line with previous geometric decomposition work, we desire a representation with as few primitives as possible. In this work, we show that our method can vary the number of primitives per-scene by initially starting with a fixed number, and allowing a post-process to find and remove unnecessary convexes.
\end{enumerate}

Our contributions are:
\begin{enumerate}
\item Our primitive decomposition method for indoor scenes is a simple procedure that works well on established and novel metrics on the benchmark NYUv2 dataset. In particular, we are the first, to our knowledge, to evaluate primitive decompositions using standard metrics for depth, normal, and segmentation.
\item We show our mixed strategy offers drastic advantages over either descent or regression methods in isolation. The key insight is that convex fitting to noisy indoor scenes is extremely difficult via pure optimization, but very good start points can be generated via regression.
\item We are the first to use segmentation labels to drive primitive generation.
\end{enumerate}

\begin{figure}
  \centering
  \begin{tabular}{@{}c@{}}
    \includegraphics[width=.90\linewidth]{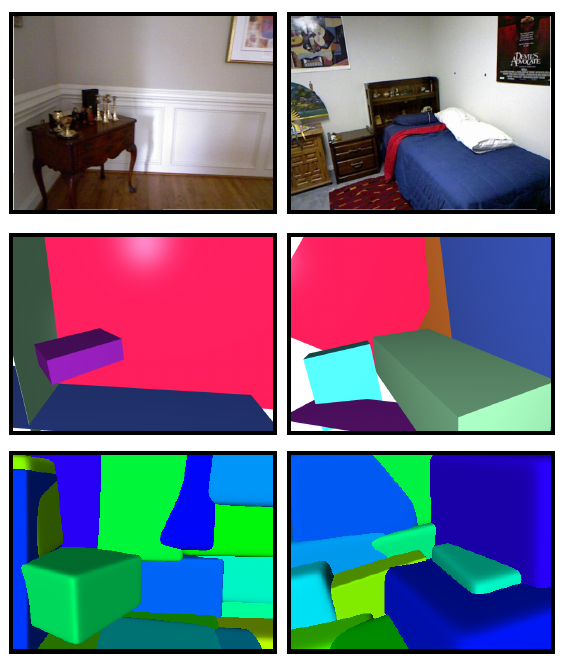}
  \end{tabular}
      \caption{(\textit{top row}) RGB image input, (\textit{middle row}) cuboid decomposition predicted by \cite{kluger2021cuboids}, (\textit{bottom}) our decomposition. Qualitative comparison of our method against previous work. Our method achieves better coverage of the input image (notice white patches in \cite{kluger2021cuboids}), enabling direct comparison of the resulting depth and normals against the GT using established techniques instead of hand-crafted metrics.}
    \label{fig:teas2}
    \vspace{-4mm}
\end{figure}

\begin{figure*}[h!]
    \centering
  \includegraphics[width=0.9\textwidth]{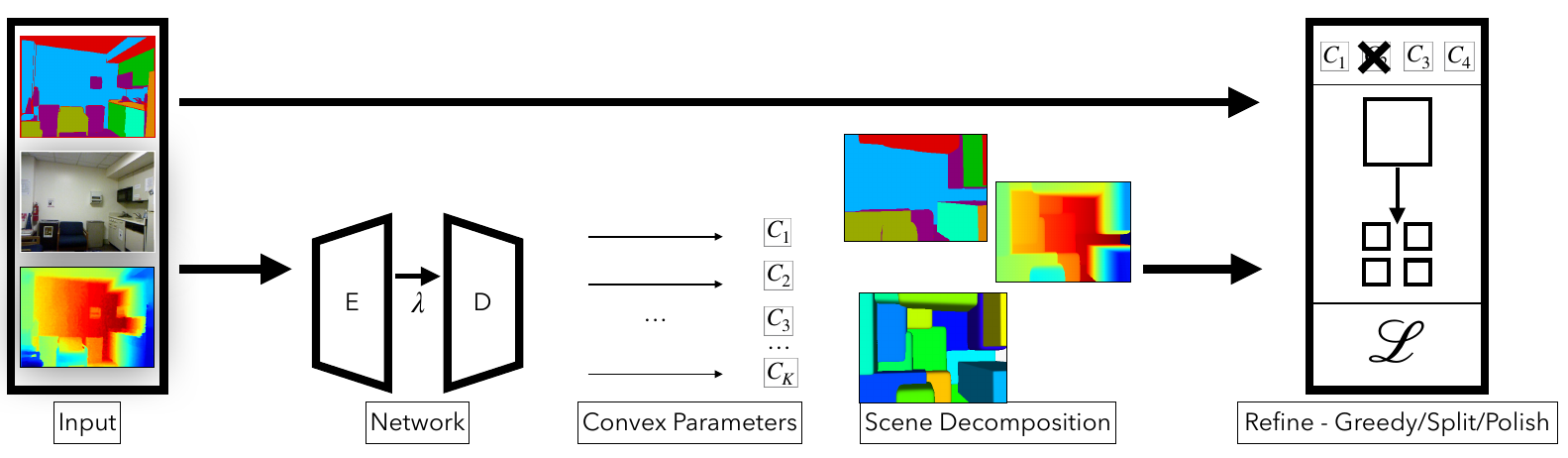}
  \caption{An overview of our inference procedure. An RGBD image and segmentation map is the input to an encoder-decoder. A series of losses supervise the training process, requiring samples labeled as ``inside" or ``outside." The network tries to adjust the convexes to classify the points correctly. The network output can be optionally refined by greedily removing primitives that do not increase the loss, splitting convexes (effectively trading off parsimony for granularity) and finally directly optimizing the convex parameters with respect to the losses. This refinement procedure accepts a segmentation map and depth map (either GT or inferred e.g. ~\cite{Ranftl2020,Ranftl2021,cao2021shapeconv}). Note that the network is unaware of the refinement process during training - it just tries to predict the best decomposition possible via the losses. The final decomposition can be visualized by ray marching from the original viewpoint and labeling each convex with a different color. Unlike previous methods, we can evaluate our approach against the original depth map by using traditional depth and normal error metrics.}
  \label{fig:flow}
  \vspace{-4mm}
\end{figure*}
\section{Related Work}
\label{sec:related}

There is a huge early literature on decomposing scenes or
objects into primitives for computer vision. Roberts worked with scenes and actual blocks ~\cite{roberts}; Binford with
generalized cylinder object descriptions~\cite{binford71}; and, in human vision, Biederman with
geon descriptions of objects~\cite{biederman}.
A representation in terms of primitives offers {\em parsimonious abstraction} (complex
objects could be handled with simple primitives; an idea of broad scope~\cite{Chen2019BSPNetGC}) and {\em natural
 segmentation} (each primitive is a part~\cite{biederman,binford71,lego}). 
Desirable primitives should be easily imputed from image data~\cite{nevatia77,shgc}, and allow
simplified geometric reasoning~\cite{hebertponce}. Generally, fitting was by choosing a
set of primitives that minimizes some cost function ({\bf descent}).

There has been much focus on individual objects (as opposed to scenes). Powerful neural methods offer the prospect of
learning to predict the right set of primitives from data -- a {\bf regression} method. Such methods can avoid local
minima by predicting solutions for test data that are ``like'' those that worked for training data. Tulsiani {\em et
al.} demonstrate a learned procedure that can parse 3D shapes into 
cuboids, trained without ground truth segmentations~\cite{abstractionTulsiani17}. Zou {\em et al.} demonstrate parses
using a recurrent architecture \cite{Zou_2018_CVPR}. Liu {\em et al.} produce detailed reconstructions of objects in
indoor scenes, but do not attempt parsimonious abstraction~\cite{liu2022towards}. There is alarming evidence that 3D reconstruction networks
rely on object identity~\cite{Tatarchenko2019WhatDS}. Deng {\em et al.} (CVXNet) recover
representations of objects as a union of convexes from point cloud and image data, again without ground truth
segmentations~\cite{deng2020cvxnet}. Individual convexes appear to map rather well to parts. An early variant of CVXNet
can recover 3D representations of poses from single images, with reasonable parses into parts~\cite{deng2019cerberus}.
Meshes can be decomposed into near convex primitives, by a form of search~\cite{weiacd}.
Part decompositions have attractive editability~\cite{spaghetti}. Regression methods face some difficulty producing
different numbers of primitives per scene (CVXNet uses a fixed number;~\cite{abstractionTulsiani17} predicts the
probability a primitive is present; one also might use Gumbel softmax~\cite{jang2017categorical}).
Primitives that have been explored include: cuboids~\cite{Calderon2017BoundingPF,Gadelha2020LearningGM,Mo2019StructureNetHG,abstractionTulsiani17,Roberts2021LSDStructureNetML,Smirnov2019DeepPS,Sun2019LearningAH,Kluger2021CuboidsRL};
superquadrics~\cite{Barr1981SuperquadricsAA,Jakli2000SegmentationAR,Paschalidou2019SuperquadricsRL}; planes~\cite{Chen2019BSPNetGC,Liu2018PlaneRCNN3P}; and generalized cylinders~\cite{nevatia77, Zou20173DPRNNGS, Li2018SupervisedFO}.
There is a recent review in~\cite{fureview}. 

Another important case is parsing outdoor scenes. Hoiem {\em et al} parse into vertical and horizontal
surfaces~\cite{hoiem:2005:popup,hoiemijcv2007}; Gupta {\em et al} demonstrate a parse into blocks~\cite{s:gupta10}. Yet
another is indoor scenes. From images of indoor scenes, one can recover: a room as a cuboid~\cite{hedauiccv2009}; beds
and some furniture as boxes~\cite{hedaueccv2010}; free space~\cite{hedaucvpr2012}; and plane layouts
~\cite{stekovic2020general, liu2018planenet}. If RGBD is available, one can recover layout in
detail~\cite{Zou_2017_ICCV}. Patch-like primitives can be imputed from data~\cite{Fouhey13}. Jiang demonstrates parsing
RGBD images into primitives by solving a 0-1 quadratic program~\cite{jiang2014finding}. Like that work, we evaluate
segmentation by primitives (see~\cite{jiang2014finding}, p. 12), but we use original NYUv2 labels instead of the
drastically simplified ones in the prior work. Also, our primitives are truly convex. Most similar to our
work is that of Kluger {\em et al}, who identify cuboids sequentially with a RANSAC-like greedy
algorithm~\cite{kluger2021cuboids}. In contrast, our network parses the entire depth map in one go, and the primitives
are subsequently refined.

The success of a descent method depends critically on the start point, typically dealt with using
greedy algorithms (rooted in~\cite{ransac}; note the prevalence of RANSAC in a recent review~\cite{Kang2020ARO});
randomized search~\cite{Ramamonjisoa2022MonteBoxFinderDA,Hampali2021MonteCS}; or multiple starts.
Regression methods must minimize loss over all training data, so at inference
time do not necessarily produce the best representation for the particular scene. The prediction is biased by the need to get other scenes right, too. To manage this difficulty, we use a mixed reconstruction strategy -- first, predict primitives using a network,
then polish using descent combined with backward selection.

It is usual that segmentations are a by-product of fitting primitives, and that regression methods are learned
without any segmentation information. In contrast, our method uses semantic segmentation results to both predict
and polish primitives. We do so because we predict primitives from images of real scenes (rather than renderings of
isolated objects), and reliable methods for semantic segmentation of scenes are available. We know of no other method
that exploits segmentation results in this way.

\begin{figure*}[t!]
\centering
  \includegraphics[width=0.9\textwidth]{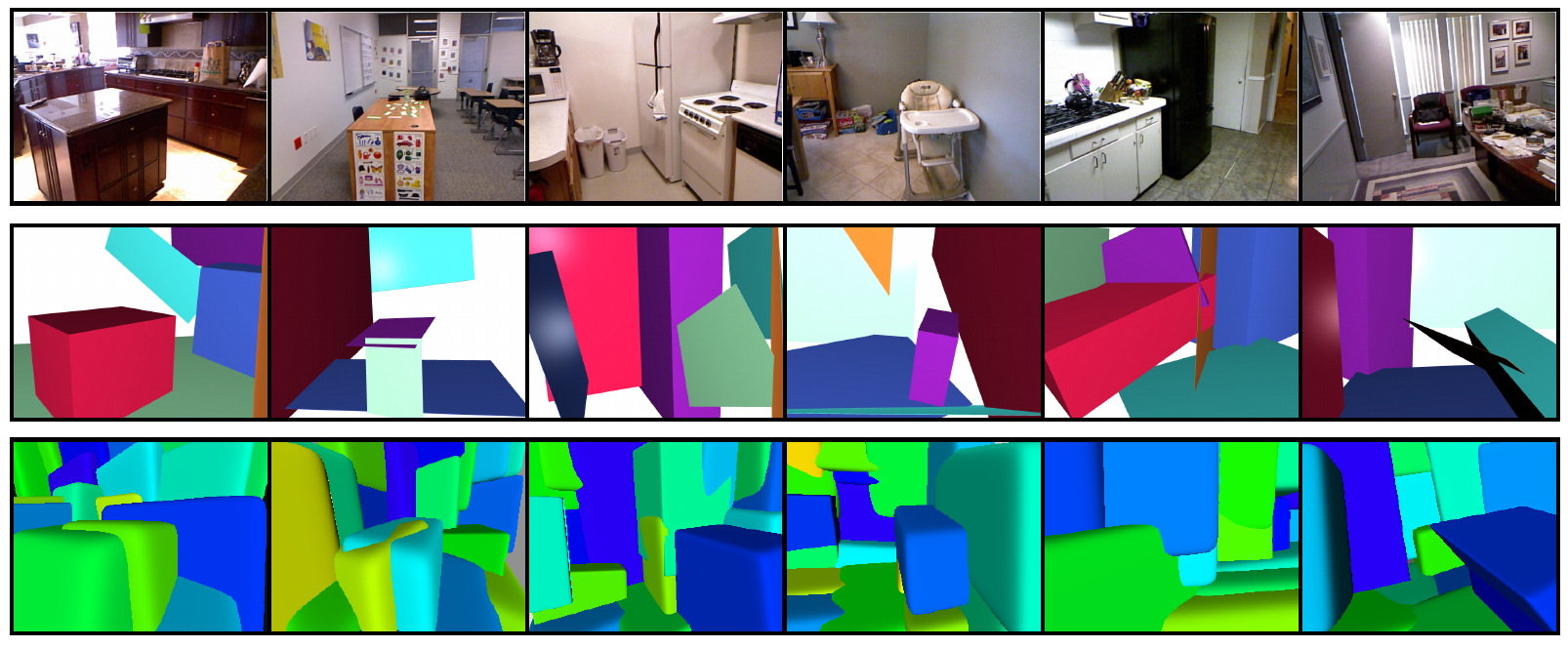}
        \caption{Results from our convex decomposition method (third row) as compared with prior work from \cite{kluger2021cuboids} (second row), which does not model the whole input. Using our data generation strategy and hybrid method, our primitives obtain better coverage. Our refinement and pruning process started from a fixed number of convexes (24 in this example) and was able to represent these scenes with an average of 14 convexes.}
        \label{fig:six}
 \vspace{-4mm}       
\end{figure*}

\section{Method - Convex Decomposition}

At runtime, we seek a collection of primitives that minimizes a set of losses \ref{sec:losses}. We will obtain this by (a) passing inputs into a network to predict a start point then (b) polishing these primitives. The network is trained to produce a fixed length set of primitives that has a small loss \ref{sec:learning}. Polishing uses a combination of descent and greedy strategies \ref{sec:polishing}. We summarize our procedure in Fig. \ref{fig:flow}.

\subsection{Losses}
\label{sec:losses}
Our losses use samples from depth \ref{sec:sampling}. We begin with the losses and hyperparameters of \cite{deng2020cvxnet}. A signed distance function of each point $\Phi(\textbf{x})$ against every halfplane is defined using logsumexp to facilitate smooth learning. An indicator function based on sigmoid outputs a continuous classification score per-point between $[0,1]$ effectively indicating whether a sample is outside all convexes or inside at least one. The primary loss is a sample loss that encourages the convexes to classify each sample correctly. We found the remaining auxiliary losses in the paper to be beneficial and adopted them with default hyperparameters. Additionally, in our early experiments we noticed long and thin convexes generated occasionally during training, whereas the types of convexes we're interested in tend to be cuboidal. Thus we modify the \textbf{unique parametrization loss} to penalize small offsets:

\begin{equation}
\mathcal{L}_{unique} = \frac{1}{H}\sum_h ||d_h||^2 + \frac{1}{H}\sum_h ||1/d_h||^2 
\end{equation}
\noindent where $H=6$ is the number of halfplanes per convex.

\textbf{Manhattan world losses} We introduce three auxiliary losses to help orient the convexes in a more structured way. We have the network predict 9 additional parameters $\mathcal{M} \in \mathbb{R}^{3\times 3}$ encoding a Manhattan world, and encourage orthonormality with the loss:
\begin{equation}
\mathcal{L}_{ortho} = \frac{1}{9}\sum(\mathcal{I}-\mathcal{M}'\mathcal{M})^2
\end{equation}

\noindent where $\mathcal{M}'$ is the transpose, and $\mathcal{I}$ is the identity. The weight of this loss is $10$.

Next, we must introduce an alignment loss that encourages the faces of each convex to follow the Manhattan world (which forces us to fix the number of planes per convex to 6 for simplicity). Let $\mathcal{N}$ be the $3\times 6$ matrix containing the predicted normals of a given convex as its columns. Let $\mathcal{W}=[\mathcal{M}; -\mathcal{M}]$ be the $3\times 6$ matrix capturing the Manhattan world basis and its negative. The second auxiliary loss is given by maximizing the element-wise dot product between $\mathcal{N}$ and $\mathcal{W}$:
\begin{equation}
\mathcal{L}_{align} = 1 - \frac{1}{6} \sum \mathcal{N}\cdot \mathcal{W}
\end{equation}

\noindent We normalize the Manhattan world vectors in $M$ before applying $\mathcal{L}_{align}$, and set its weight to $1$. This loss effectively enforces an order for the normals within each convex.

Finally, we establish an explicit ordering of the convexes themselves, encouraging the network to generate them in order of volume. Since the predicted convexes are approximately cuboids, the volume of a convex can be estimated with $(d_0+d_3) * (d_1+d_4) * (d_2+d_5)$ where $d_i$ is the offset for halfplane $i$. Given a vector $\mathcal{V} \in \mathbb{R}^{K}$ of volumes where there are $K$ convexes at training time, we can encourage the network generate convexes in order of volume:

\begin{equation}
\mathcal{L}_{vol} = \frac{1}{K} \sum{}ReLU(\mathcal{V}[1:] - \mathcal{V}[0:-1])
\end{equation}

\noindent The weight of this loss is $1$. In practice, we observed that the first (and largest convex) was almost always the floor. Downstream applications may benefit from having the convexes presented sorted by size. As we show in our ablation Table~\ref{tab:err_abl}, the volume loss has an approx. neutral effect on the error metrics but did improve parsimony.

We set the number of convexes $K = 24$. With more convexes, structures that could be explained by one convex may be explained by multiple potentially overlapping convexes in some scenes, which we notice in practice. However, segmentation accuracy generally improves with more parts due to the increased granularity.

\indent\textbf{Segmentation Loss} We construct a loss to encourage the convexes to respect segmentation boundaries.  Ideally, each face of each convex (or, alternatively, each convex) spans one label category in the image.  To compute this loss, we obtain a smoothed distribution of labels spanned by a face (resp. convex), then penalize the entropy of the distribution. This loss penalizes convexes that span many distinct labels without attending to label semantics.  We use the standard 40 NYUv2 labels, with one background label.

To compute the loss for convexes, write ${\cal C}$ for the $K \times N$ matrix representing the indicator function response for each inside surface sample for each convex,
and ${\cal L}$ for the $N \times 41$ one-hot encoding of the segment label for each sample. Then each row of ${\cal C} {\cal L}$ represents the count of labels within each convex, weighted so that a sample that is shared between
convexes shares its label. We renormalize each row to a distribution, and compute the sum of entropies. To compute the loss for convex faces we obtain ${\cal C}_f$, $(6K) \times N$ matrix, where each row of ${\cal C}_f$ corresponds
to a face.  For the row corresponding to the $i$'th face of the $k$'th convex, we use the $k$'th row of ${\cal C}$ masked to zero if the closest face to the sample is not $i$. Then each row of ${\cal C}_f {\cal L}$ represents the count of labels spanned by each face, weighted so that a sample that is shared between faces shares its label. We renormalize each row to a distribution, and compute the sum of entropies.

\begin{equation}
\mathcal{L}_{entropy} = \frac{1}{6K} \sum entropy({\cal C}_f {\cal L})
\label{eq:entropy}
\end{equation}

We train with GT segmentation labels from NYUv2. During refinement, we can either use a GT segmentation map if available, or an inferred map (we use \cite{cao2021shapeconv}). We experimentally weight $\mathcal{L}_{entropy}$ to 1.

\begin{table*}[t!]
\begin{center}
\resizebox{\textwidth}{!}{%
\begin{tabular}{||c | c c c c c c|| c c || c c c c c || c ||} 
 \hline
  Cfg. & Prune & Refine  & $Depth_{GT}$ & $Seg_{GT}$ & $n_{init}$ & $n_{used}$ & AbsRel$\downarrow$ & RMSE$\downarrow$  & Mean$\downarrow$ & Median$\downarrow$ & $11.25^{\circ}\uparrow$ & $22.5^{\circ}\uparrow$ & $30^{\circ}\uparrow$ & $Seg_{Acc}\uparrow$ \\ 
 \hline\hline
noInit & $\times$ & $\checkmark$ & $\checkmark$ & $\checkmark$ & 24 & 24.0 & 0.447 & 1.604 & 81.286 & 80.566 & 0.021 & 0.061 & 0.096 & 0.359 \\
\hline
noSeg & $\times$ & $\times$ & $\times$ & $\times$ & 24 & 24.0 & 0.179 & 0.664 & 41.687 & 38.224 & 0.115 & 0.291 & 0.391 & \textbf{0.626} \\
\hline
withSeg & $\times$ & $\times$ & $\times$ & $\times$ & 24 & 24.0 & 0.310 & 1.231 & 52.305 & 48.846 & 0.088 & 0.231 & 0.317 & 0.528 \\
\hline
A & $\times$ & $\checkmark$ & $\times$ & $\times$ & 24 & 24.0 & 0.163 & 0.679 & 40.692 & 35.697 & 0.124 & 0.313 & 0.424 & 0.623 \\
\hline
B & $\times$ & $\checkmark$ & $\times$ & $\times$ & 24 & 24.0 & 0.166 & 0.696 & 41.019 & 35.964 & 0.122 & 0.310 & 0.421 & 0.623 \\
\hline
C & $\checkmark$ & $\checkmark$ & $\times$ & $\times$ & 24 & 14.4 & \textbf{0.144} & \textbf{0.603} & \textbf{38.235} & \textbf{33.621} & \textbf{0.133} & \textbf{0.335} & \textbf{0.451} & 0.615 \\
\hline\hline
D & $\checkmark$ & $\checkmark$ & $\checkmark$ & $\times$ & 24 & 14.0 & \textit{0.098} & \textit{0.513} & 37.361 & 32.402 & \textit{0.144} & \textit{0.353} & \textit{0.469} & \textit{0.619} \\
 \hline
E & $\checkmark$ & $\checkmark$ & $\checkmark$ & $\checkmark$ & 24 & 13.9 & \textit{0.098} & 0.514 & \textit{37.355} & \textit{32.395} & \textit{0.144} & \textit{0.353} & \textit{0.469} & 0.618 \\
\hline\hline
ref & - & - & $\checkmark$ & $\checkmark$ & - & - & 0.110 & 0.357 & 14.9 & 7.5 & 0.622 & 0.793 & 0.852 & 0.719 \\
 \hline\hline
\end{tabular}}
\vspace{0.1cm}
\caption{Ablation study with different configurations on 654 NYUv2 test images, all error metrics evaluated against ground truth. P refers to whether pruning is applied during refinement, R refers to whether refinement optimization steps are taken, $D_{GT}$ and $Seg_{GT}$ refer to whether GT depth and segmentation were available during refinement (and inferred by pretrained networks if not).  $n_{start}$ refers to the number of parts available at training time $n_{used}$ refers to the mean number of convexes remaining post-refinement. Unlike previous work, we can evaluate directly against ground truth depth and normals. We evaluate depth error metrics (AbsRel/RMSE) \cite{Ranftl2021}, normal error metrics (subsequent 5 columns) \cite{wang2015designing}, and pixel-wise segmentation accuracy (last column) on the 654 NYUv2 test images previous work considers \cite{lee2019big,kluger2021cuboids}. Segmentation accuracy assumes that we label each convex with the most common label within its boundary from the ground truth segmentation before computing mean pixel-wise accuracy. \textbf{noInit} Polishing convexes from a random start point yields very poor results. Since we cannot trust refinement with an arbitrary start point, we train a neural network (on the NYUv2 795 train split). \textbf{noSeg} The convexes predicted by a neural network without refinement offer a dramatic improvement over directly optimizing from a random start point. \textbf{withSeg} When training with the segmentation loss, all error metrics are worse. We're not sure why this is, but as we show below, all error metrics significantly improve when we add refinement. \textbf{A} We refine the network prediction with all losses except entropy (for segmentation). Depth map for refinement is inferred by \cite{Ranftl2021}. We notice further improvements in depth and normal metrics, but slightly worse segmentation accuracy. \textbf{B} Same as A, but we refine with the segmentation loss, where segmentation maps are inferred by \cite{cao2021shapeconv}. Overall, introducing the segmentation loss has a neutral effect across the board. \textbf{C} We allow pruning unused convexes in the refinement process via backward selection (note how $n_{used}$ drops). Observe how depth and normal metrics improve, but segmentation is slightly worse due to there existing fewer convexes. \textbf{D} Same as C, but allow GT depth maps during refinement instead of inferred. We see a drastic improvement in depth and normal error metrics. \textbf{E} we allow GT segmentation maps during refinement instead of inferred. This has an overall neutral effect on the quality as compared with D. D and E rely on oracles and the best metrics for oracle systems are italicized. The best non-oracle cfg. was C, and the best metrics are bolded. The last row \textbf{ref} shows the reference error of our pretrained depth estimation network, a recent normal estimation work, and our pretrained segmentation network \cite{Ranftl2021,bae2021estimating,cao2021shapeconv}. Our primitive decompositions using inferred depth/seg in C show depth and segmentation metrics that are only a bit worse than ref., despite the challenging task of representing the scene with just a few primitives. But the normals show a significant gap - this could be due to the presence of curved or complex geometry that can achieve good depth error with just a simple convex, but cannot capture the intricacies in the normals. Further ablations in supplement.}

\label{table:err}
\end{center}
\vspace{-6mm}
\end{table*}

\subsection{Polishing} 
\label{sec:polishing}

Our network produces reasonable decompositions. However,
it does not produce the best possible fit for a particular input. The network is a mapping that produces an output with a good loss value {\em on average}. Unlike the usual case, we can provide all necessary information to evaluate losses {\em at inference time}, and so we can polish the network output for a given input. Thus to improve our primitives, we apply a descent procedure on the network predictions using the original loss terms applied to a very large number of samples from the test-time input, refining our predicted primitives significantly. 

The number of samples per scene at training time is limited (18K per image); the refinement process allows using a large number of samples (we use 250K) during optimization because the batch size is only 1. We refine for 500 iterations with SGD optimizer, learning rate $0.01$, adding about 40 seconds to the inference time. 

Furthermore, in our experimentation we have not found existing methods to control the number of primitives particularly effective for this problem~\cite{abstractionTulsiani17}. Thus in this work we use simple backward selection to remove primitives. For each primitive, we evaluate the loss with that primitive removed. If the loss does not go up by more than $\epsilon_{prune} = 0.001$ then we remove it and continue searching. In practice, this removal procedure requires $K$ additional forward passes through the network but helps find a more parsimonious representation of the input. 

We remind the reader that the network input is an RGBD image (both training and test time). When applying the optional segmentation loss eqn.~\ref{eq:entropy}, a segmentation map is required. At test time, the network still requires an RGBD image to obtain an initial set of primitives; if applying polishing, a depth map is required and segmentation map is optional. At test time, ``oracle'' refers to GT depth and segmentation being available, whereas ``non-oracle'' refers to depth and segmentation maps inferred by off-the-shelf depth/segmentation networks.

Finally, before applying this polishing, we can optionally increase the number of primitives by splitting each convex into 8 pieces. For parallelepipeds, the child convexes can inherit the parent's normals, child offsets are half of the parent's, and new translations can be computed by taking the midpoint of the original translation coordinate and each of the 8 corners. We focus our experimentation on 0 or 1 splits, as the number of parts grows very quickly. Future work may consider only splitting convexes that show high error with respect to the provided depth or segmentation. 

\subsection{Sampling} 
\label{sec:sampling}
The philosophy of our sampling strategy is to encourage placement of convexes to overlap objects in the scene, and discourage placement of convexes where we do not want them. Our losses require samples that can be obtained from the input depth map and camera calibration parameters translating each pixel to a world space 3D coordinate. Given those world space coordinates, we then re-scale the coordinates (per depth map) to lie in a range amenable to training neural networks (approx. a unit cube). Because we work with 240x320 pixel depth maps (4:3 aspect ratio), we rescale X coordinates to lie between $[-\frac{2}{3},\frac{2}{3}]$; Y coordinates between $[-0.5, 0.5]$ and Z coordinates between [0,0.8]. Downstream we need the parameters used to transform a depth map back in the original scale to ray march the predicted decomposition. 

\textbf{Surface samples} From there we need to generate two classes of samples: ``free space" and ``surface." Each sample is a 4-tuple combining the 3D normalized coordinate and a binary label indicating whether the sample is inside the desired volume (1) or outside (0). Inside surface samples are generated by taking the normalized depth coordinates and adding a small constant to Z (we use $\epsilon_{surf}=0.03$, obtained experimentally) resulting in the tuple $(X,Y,Z+\epsilon_{surf},1)$. Analogously, outside surface samples are given by $(X,Y,Z-\epsilon_{surf},0)$. 

\textbf{Free space samples} We can generate free space samples on the fly as follows. Cast orthographic rays along the Z dimension through each pixel and sample along that ray. If the depth value of the random sample is less than the $Z$ value from the depth at that pixel, that sample is classified as outside. If the depth of the random sample lies between $[Z, Z+t]$ (where $t$ is a thickness parameter we experimentally set to $0.1$) then we classify the sample as inside. The $Z$-coordinate of a random free space sample for pixel $(i,j)$ ranges from $[-m_d,Z(i,j)+t]$ where we set $m_d=-0.1$ and $Z(i,j)$ is the re-scaled depth value at pixel $(i,j)$. To avoid convexes deviating far from the inside samples, we also construct a shell of "outside" free space samples around all of the aforementioned samples. This shell can be thought of as the difference of two cubes with identical center but different edge length. We let the thickness of this shell be $0.3$ and let it start at $\pm 1.2$ along the $x$-direction, $\pm 1.0$ along the $y$-direction, and $[-0.5, 1.3]$ along the $z$-direction, leaving a min. 0.4 margin between any inside sample and these outside shell samples along all axes. Thus during the training process the convexes have some volume to move freely without any samples encouraging convexes to exist or not exist in this "gap" and they can focus on getting samples near the depth boundary classified correctly.

\subsection{Learning}
\label{sec:learning}
As Table \ref{table:err} indicates, polishing our loss from a random start produces very weak results. Thus training a network to produce a start point is essential.

\textbf{Data} Our training procedure requires depth maps with known camera calibration parameters. Like previous scene parsing work, we focus on NYUv2 \cite{Silberman:ECCV12}, containing 1449 RGBD images. We use the standard 795-654 train-test split. While 795 images is usually quite low for training neural networks, we show that the decomposition predicted by the network is actually a good start point for subsequent refinement that achieves a very good fit.

\textbf{Network} We use the architecture of \cite{deng2020cvxnet} consisting of a ResNet18 encoder that accepts an RGBD image and a decoder consisting of 3 fully-connected layers, 1024 neurons each. The output of the decoder consists of parameters representing the translations and blending weight of each convex as well as the normal and offset of each halfplane. The number of halfplanes and convexes is fixed during training time, which determines the number of output parameters. By forcing each predicted convex to be a parallelepiped, we only need to predict three normals per convex instead of six, which eased training in our experiments.

\textbf{Training Details} We implement our method with TensorFlow and train on one NVIDIA A40 GPU. We use Adam Optimizer and learning rate 0.0001. We set our minibatch size to 32 images (requiring 2.5 hours for 20k iterations) and randomly sample 6k free space and 12k surface samples per image in each element of the minibatch. 10\% of the free space samples are taken from the outside shell. Within the sample loss, we also observed slightly better quality and more stable training by annealing the relative weight of free space samples vs. surface samples, starting from a $0.9:0.1$ split and saturating to $0.5:0.5$ midway through training. Intuitively, this allows the network to focus on big picture features of the geometric layout early on, and then adjust fine-scale details of the surface near the end.

\begin{figure}[t!]
\centering
\begin{tabular}{@{}c@{}}
  \includegraphics[width=0.95\linewidth]{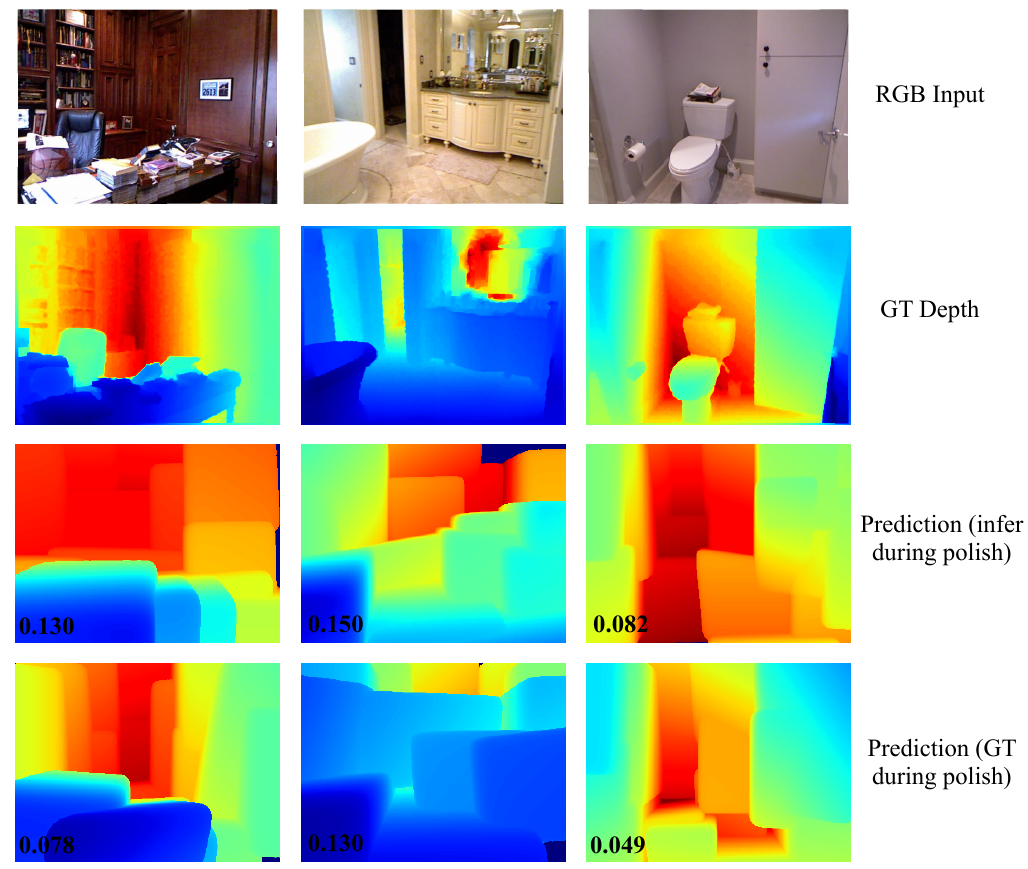} 
  \end{tabular}
  \caption{Qualitative evaluation of our method on depth. These are random test images not seen during the training process. The third row shows having only inferred depth and segmentation during refinement, whereas the last row shows having GT depth/seg available during refinement. AbsRel depth superimposed over predictions. Fairly good approximations of depth can be generated with just a few convexes - 24 at train time, and about 14 remain after refinement. Only a small penalty in quality is observed by not having GT depth/seg available.}
\label{fig:depth}
\vspace{-1mm}
\end{figure}

\begin{figure}[t!]
\centering
\begin{tabular}{@{}c@{}}
  \includegraphics[width=0.95\linewidth]{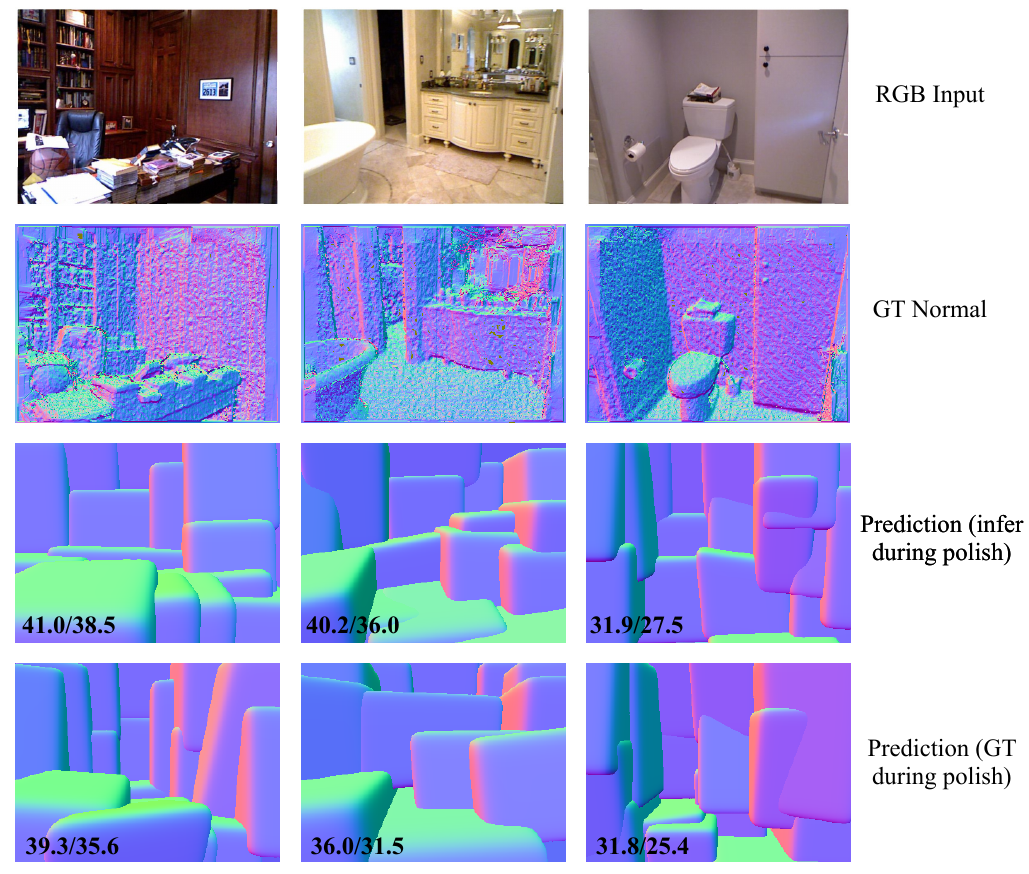} 
  \end{tabular}
  \caption{Qualitative evaluation of our method on normals. These are random test images not seen during the training process. The third row shows having only inferred depth and segmentation during refinement, whereas the last row shows having GT depth/seg available during refinement. Mean/median normal error in degrees superimposed over predictions. Fairly good approximations of normals can be generated with just a few convexes - 24 at train time, and about 14 remain after refinement.}
  \label{fig:norm}
  \vspace{-4mm}
\end{figure}

\begin{figure}[t!]
\centering
\begin{tabular}{@{}c@{}}
  \includegraphics[width=0.95\linewidth]{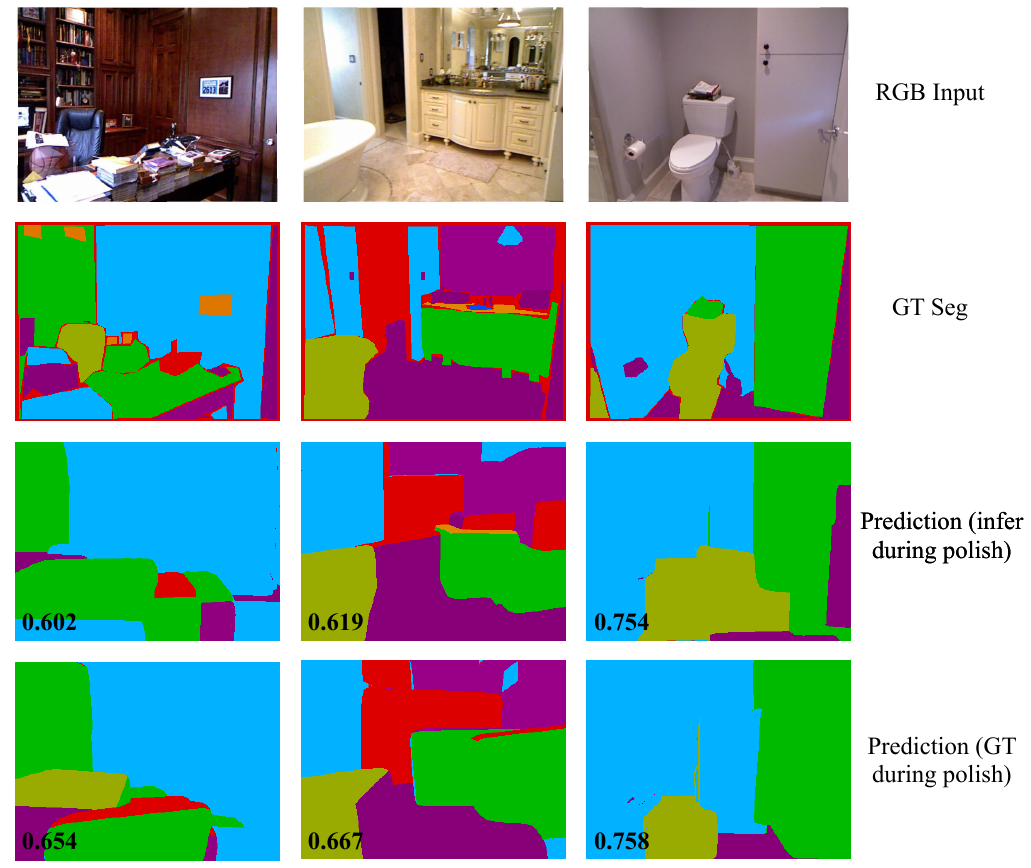} 
  \end{tabular}
  \caption{Qualitative evaluation of our method on segmentation. These are random test images not seen during the training process. The third row shows having only inferred depth and segmentation during refinement, whereas the last row shows having GT depth/seg available during refinement. Mean per-pixel segmentation accuracy is superimposed over predictions. Fairly good approximations of segmentation can be generated with just a few convexes - 24 at train time, and about 14 remain after refinement.}
    \label{fig:seg}
    \vspace{-4mm}
\end{figure}

\begin{figure*}[t!]
\centering
  \includegraphics[width=0.95\textwidth]{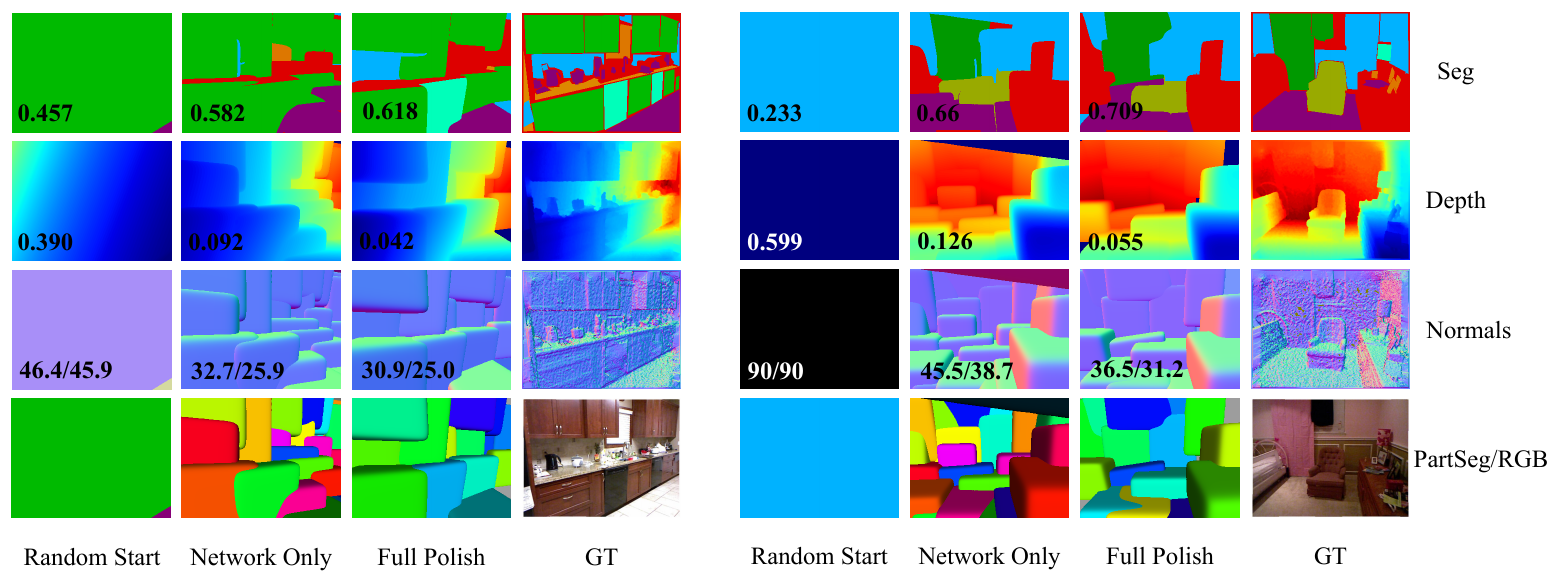}
  \caption{Qualitative ablation study of our method. Two random test images never seen by the network are shown. The first row shows segmentations, the second depth, the third normals, and the fourth shows convex segmentations for ease of visualization (as well as the RGB image input in the fourth and eight columns). \textbf{Random Start} Randomly initializing convexes followed by polishing does not yield results resembling the input. This is an extremely difficult optimization problem. \textbf{Network Only} By training a neural network to predict primitives, we get fairly good results, even without subsequent polishing. \textbf{Full Polish} When we polish the result by performing backward selection to remove extraneous convexes and perform additional optimization steps on the convex parameters, we can improve the overall fit. Notice how polishing removed an extraneous primitive at the top of the scene in column 6. Mean segmentation accuracy, depth AbsRel, and normal mean/median error are overlaid.}
  \label{fig:ablation}
  \vspace{-2mm}
\end{figure*}


\section{Experiments \& Evaluation}
\label{sec:exper}
We investigate several design choices in our pipeline. The primary finding is that our network alone (regression) produces reasonable but not great convex decompositions. Polishing alone (descent) generates completely unusable primitives. But allowing network input as initialization of our refinement procedure results in our best quality. Our results remain strong even when ground truth depths and segmentations are not available during refinement and must be inferred by other networks. In the remainder of this section, we show that our mixed procedure produces good depth, normals, and segmentation on our test dataset.

\textbf{Quantitative evaluation of normals and depth}: Given parameters of a collection of convexes for a given test image, we can ray march from the original viewpoint to obtain a depth map, part segmentation, and normals. Our method involves ray marching and interval halving at the first intersection point. In Table \ref{table:err} we use the standard depth error metrics to evaluate the depth associated with the convex decomposition against the input depth map, evaluating on the 654 NYUv2 test images from \cite{lee2019big}. Although we train with samples that are re-normalized to a small cube near the origin, we transform the predicted depth map back to world space coordinates when computing depth error metrics.

Further, in Table \ref{table:err} we also use an existing suite of error metrics for evaluating normals of the predicted convexes \cite{wang2015designing}. We can compute the normals from the input depth map by first extracting per-pixel 3D coordinates $(X_{ij},Y_{ij},Z_{ij})$. Let $Z_x$ and $Z_y$ be image gradients of the depth map in the $x$ and $y$ directions respectively, and $X_x$ and $Y_y$ be the image gradients of the $X$ and $Y$ coordinates in $x$ and $y$ directions respectively. The final normal per pixel is given by normalizing: $(\frac{-Z_x}{Y_y},\frac{-Z_y}{X_x},1)$. 

To obtain normals from the collection of convexes, we can compute the gradient of the implicit function that defines it and evaluate it per-pixel at the first ray-surface intersection point for each pixel: 

\begin{equation}
\nabla\Phi(\textbf{x}) = \frac{\delta \sum_h \textbf{n}_h e^{\delta * (\textbf{n}_h \cdot\textbf{x} + d_h)}}{\sum_h e^{\delta * (\textbf{n}_h \cdot\textbf{x} + d_h)}}
\end{equation}

\noindent Note that sums over $h$ iterate over all the halfplanes associated with the convex that the ray hit first. 

Our error metrics strongly indicate that polishing alone produces unusable results, network alone produces better results that often resemble the input, and our very best results come from a mixed method that uses the network prediction as a start point for subsequent refining. Our pruning procedure shows that we can improve depth and normal error metrics while reducing the number of parts. However, our ablations showed that our entropy-based segmentation loss yielded a neutral effect on the segmentation quality. That loss did help in toy models we tried, so we leave it in the paper to possibly help future research.

We compute the Chamfer-L1 distance of the predicted and ground truth depth map (in meters). Our best non-oracle network (row C, Table \ref{table:err}) got a score of 1.46; our best oracle network (row E, Table~\ref{table:err}) got a score of 0.541.

{\bf Quantitative evaluation of segmentation:}  
Ideally, primitives ``explain'' objects, so each primitive should span only one object (though individual objects could consist of more than one primitive).  We can evaluate this property by comparing primitive based segmentations with ground truth for the test partition of NYUv2. We render an image where each pixel has the number of the visible primitive at that pixel. Each primitive then gets the most common label in its support.  Finally, we compute the pixel accuracy of this labeling. Note that this test is unreliable if there are many primitives (for example, one per pixel), but we have relatively few.  A high pixel accuracy rate strongly suggests that each primitive spans a single object. Our mean pixel accuracy rate for different configurations is shown in Table \ref{table:err}, last column. Our best configurations, C \& D, showed a segmentation accuracy rate of 0.618. We also show qualitative results in Fig. \ref{fig:seg}. We conclude that our primitive decomposition mostly identifies objects, but could likely be improved by having more primitives.

\textbf{Qualitative evaluation} We compare our method against previous work \cite{kluger2021cuboids} on six NYUv2 images in Fig. \ref{fig:six}. Our representation generally covers the whole depth map without leaving gaps, and finds more convexes that correspond with real-world objects. We also evaluate our method specifically looking at depth (Fig. \ref{fig:depth}), normals (Fig. \ref{fig:norm}), and segmentation (Fig. \ref{fig:seg}). Our network-initialized refinement procedure accurately predicts convexes whose depth and normals closely match the ground truth. This is true even when GT depth isn't available to the refinement. Finally, we evaluate whether our hybrid method is necessary in Fig.~\ref{fig:ablation}, showing that polishing without a network (random start) does not produce useful results due to the extremely difficult optimization problem; fair quality with running our network but no polishing; and our very best quality when running our network followed by polishing (hybrid method).

\section{Conclusion}
We presented a novel procedure to decompose complex indoor scenes into 3D primitives, relying on both regression and descent. By capturing the whole input, we used existing depth, normal, and segmentation error metrics to evaluate, unlike previous work. Further, our qualitative evaluation shows that our primitives map to real-world objects quite well, capturing more of the input than previous work. By relying on suitable pretrained networks, our experiments showed successful primitive decomposition when ground truth depth or segmentation isn't available.

One application we have had success with is generating scenes conditioned on the primitives - we can edit the geometry of the scene by simply moving a cuboid, far easier than editing say a depth map~\cite{vavilala2023blocks2world}. Future work may analyze the latent space, perhaps with applications to generative modeling or scene graph construction. Training with a large scale RGBD+Segmentation dataset could also improve the start points of our method. Further, a natural extension is generating primitives that are temporally consistent across video frames.  

Our refinement process showed it's possible to control the number of primitives and their granularity even though the start point was produced by a network trained with a fixed number of convexes. Some scenes could benefit from a convex merge process such as flat walls that are decomposed into several convexes. Such a procedure may improve qualitative results when we train with more convexes than the input requires.

\section{Acknowledgment}
This material is based upon work supported by the National Science Foundation under Grant No. 
2106825 and by gifts from Amazon and Boeing.

{\small
\bibliographystyle{ieee_fullname}
\bibliography{egbib,extrabib}
}


\setcounter{table}{1}
\begin{table*}[h]
\begin{center}
\resizebox{\textwidth}{!}{%
\begin{tabular}{||c | c c c c c c c|| c c || c c c c c || c ||} 
 \hline
  Cfg. & ManWorld & Entropy  & Split & $Depth_{GT}$ & $Seg_{GT}$ & $n_{init}$ & $n_{used}$ & AbsRel$\downarrow$ & RMSE$\downarrow$  & Mean$\downarrow$ & Median$\downarrow$ & $11.25^{\circ}\uparrow$ & $22.5^{\circ}\uparrow$ & $30^{\circ}\uparrow$ & $Seg_{Acc}\uparrow$ \\ 
 \hline\hline
12-NO & $\checkmark$ & $\checkmark$ & $\times$ & $\times$ & $\times$ & 12 & 6.5 & 0.185 & 0.735 & 40.052 & 36.009 & 0.122 & 0.303 & 0.418 & 0.500 \\
\hline
24-NO & $\checkmark$ & $\checkmark$ & $\times$ & $\times$ & $\times$ & 24 & 14.4 & 0.144 & \textbf{0.603} & \textbf{38.235} & \textbf{33.621} & \textbf{0.133} & \textbf{0.335} & \textbf{0.451} & 0.615 \\
\hline

 24-NS & $\checkmark$ & $\checkmark$ & $\checkmark$ & $\times$ & $\times$ & 24 & 25.9 & 0.183 & 0.851 & 41.512 & 37.445 & 0.107 & 0.286 & 0.396 & \textbf{0.630} \\

\hline
24-NM & $\times$ & $\checkmark$ & $\times$ & $\times$ & $\times$ & 24 & 9.5 & 0.189 & 0.716 & 43.003 & 38.914 & 0.059 & 0.223 & 0.355 & 0.536 \\
\hline
24-NE & $\checkmark$ & $\times$ & $\times$ & $\times$ & $\times$ & 24 & 16.1 & \textbf{0.143} & 0.614 & 38.311 & 33.872 & 0.132 & 0.330 & 0.445 & 0.629 \\

\hline
36-NO & $\checkmark$ & $\checkmark$ & $\times$ & $\times$ & $\times$ & 36 & 12.6 & 0.198 & 0.781 & 45.546 & 41.796 & 0.062 & 0.201 & 0.317 & 0.565 \\
\hline
48-NO & $\checkmark$ & $\checkmark$ & $\times$ & $\times$ & $\times$ & 48 & 14.6 & 0.199 & 0.833 & 43.963 & 39.802 & 0.074 & 0.267 & 0.370 & 0.584 \\
\hline
60-NO & $\checkmark$ & $\checkmark$ & $\times$ & $\times$ & $\times$ & 60 & 16.9 & 0.175 & 0.665 & 42.387 & 38.418 & 0.092 & 0.269 & 0.383 & 0.600 \\
 \hline\hline
12-O & $\checkmark$ & $\checkmark$  & $\times$ & $\checkmark$ & $\checkmark$ & 12 & 6.2 & 0.153 & 0.581 & 40.246 & 36.475 & 0.120 & 0.297 & 0.416 & 0.492 \\
\hline
 24-O & $\checkmark$ & $\checkmark$ & $\times$ & $\checkmark$ & $\checkmark$ & 24 & 13.9 & 0.098 & \textit{0.514} & \textit{37.355} & \textit{32.395} & \textit{0.144} & \textit{0.353} & \textit{0.469} & 0.618 \\

\hline
24-OS & $\checkmark$ & $\checkmark$ & $\checkmark$ & $\checkmark$ & $\checkmark$ & 24 & 27.4 & 0.143 & 0.819 & 41.235 & 36.839 & 0.111 & 0.295 & 0.407 & \textit{0.631} \\
 
\hline
 24-OM & $\times$ & $\checkmark$ & $\times$ & $\checkmark$ & $\checkmark$ & 24 & 9.8 & 0.151 & 0.580 & 43.349 & 39.117 & 0.057 & 0.218 & 0.349 & 0.568 \\

 \hline
 24-OE & $\checkmark$ & $\times$ & $\times$ & $\checkmark$ & $\checkmark$ & 24 & 15.7 & \textit{0.096} & 0.520 & 37.513 & 32.700 & 0.143 & 0.347 & 0.464 & 0.630 \\

\hline

36-O & $\checkmark$ & $\checkmark$ & $\times$ & $\checkmark$  & $\checkmark$ & 36 & 13.0 & 0.165 & 0.629 & 44.189 & 40.012 & 0.066 & 0.212 & 0.335 & 0.577 \\
\hline

48-O & $\checkmark$ & $\checkmark$ & $\times$ & $\checkmark$ & $\checkmark$ & 48 & 14.8 & 0.159 & 0.657 & 44.219 & 40.025 & 0.069 & 0.255 & 0.362 & 0.597 \\
\hline
60-O & $\checkmark$ & $\checkmark$ & $\times$ & $\checkmark$ & $\checkmark$ & 60 & 17.4 & 0.172 & 0.652 & 43.771 & 39.620 & 0.089 & 0.258 & 0.369 & 0.605 \\
\hline\hline
ref & - & - & - & - & - & - & - & 0.110 & 0.357 & 14.9 & 7.5 & 0.622 & 0.793 & 0.852 & 0.719 \\
 \hline\hline
\end{tabular}}
\vspace{0.1cm}
\caption {Additional ablations. All configs have pruning and polishing applied at the refinement stage. The eight experiments between 12-NO through 60-NO refer to not having an oracle available during refinement (i.e. the depth and segmentation are inferred by \cite{Ranftl2021,cao2021shapeconv}). The next eight configs have oracles available. The last row ref. shows the reference error of our pretrained depth estimation network, a recent normal estimation work, and our pretrained segmentation network \cite{Ranftl2021,bae2021estimating,cao2021shapeconv}. By testing the number of starting parts at training time between [12-NO,24-NO,36-NO,48-NO,60-NO] as well as [12-O,24-O,36-O,48-O,60-O] and then applying refinement, 24 parts performs best. This is a classic case of bias-variance tradeoff: too few parts biases the decomposition with insufficient capacity; too many parts results in variance problems. We also evaluate the effects of our Manhattan World losses (24-NO vs. 24-NM) and (24-O vs. 24-OM); quantitatively, these harm our error metrics across the board. Clearly, indoor scenes have objects oriented in a similar way, and enforcing that as a loss improves the quality of our decompositions. From there, we examine the effects of no entropy loss (i.e. segmentation loss) during training nor inference, comparing experiments (24-NO vs. 24-NE) and (24-O vs. 24-OE). The depth and normal error metrics are nearly identical with or without this loss, and the segmentation accuracy is slightly better WITHOUT the segmentation loss. This was slightly unexpected, but could be due to the overall noisiness and complexity of the segmentation maps preventing clean segmentation of objects. We illustrate an example in Fig. \ref{fig:entropy}. Also note how the pruning process removed more parts with the segmentation loss, and it is generally to be expected that more parts can lead to better segmentation accuracy. Finally, we examine the effects of splitting each convex into 8 equal volume pieces during refinement before applying refinement/pruning. These are examined quantitatively in configs (24-NO vs. 24-NS) and (24-O vs. 24-OS). Overall, we achieved our best segmentation score with splitting, to be expected due to the increased number of parts. However, depth and normal error metrics suffer. This is due to optimization difficulties - the optimizer struggled to improve the fit of so many parts. Qualitatively, the pruning process resulted in holes in the representation as shown in Fig.~\ref{fig:spp1}. We think that additional investigation into splitting and pruning can lead to near-arbitrary resolution convex decompositions, an exciting next step. 
}
\label{tab:err1}
\end{center}
\vspace{-6mm}
\end{table*}


\section{Additional Ablations}
\label{sec:start1}
\begin{figure*}
\centering
\begin{tabular}{@{}c@{}}
  \includegraphics[width=0.95\linewidth]{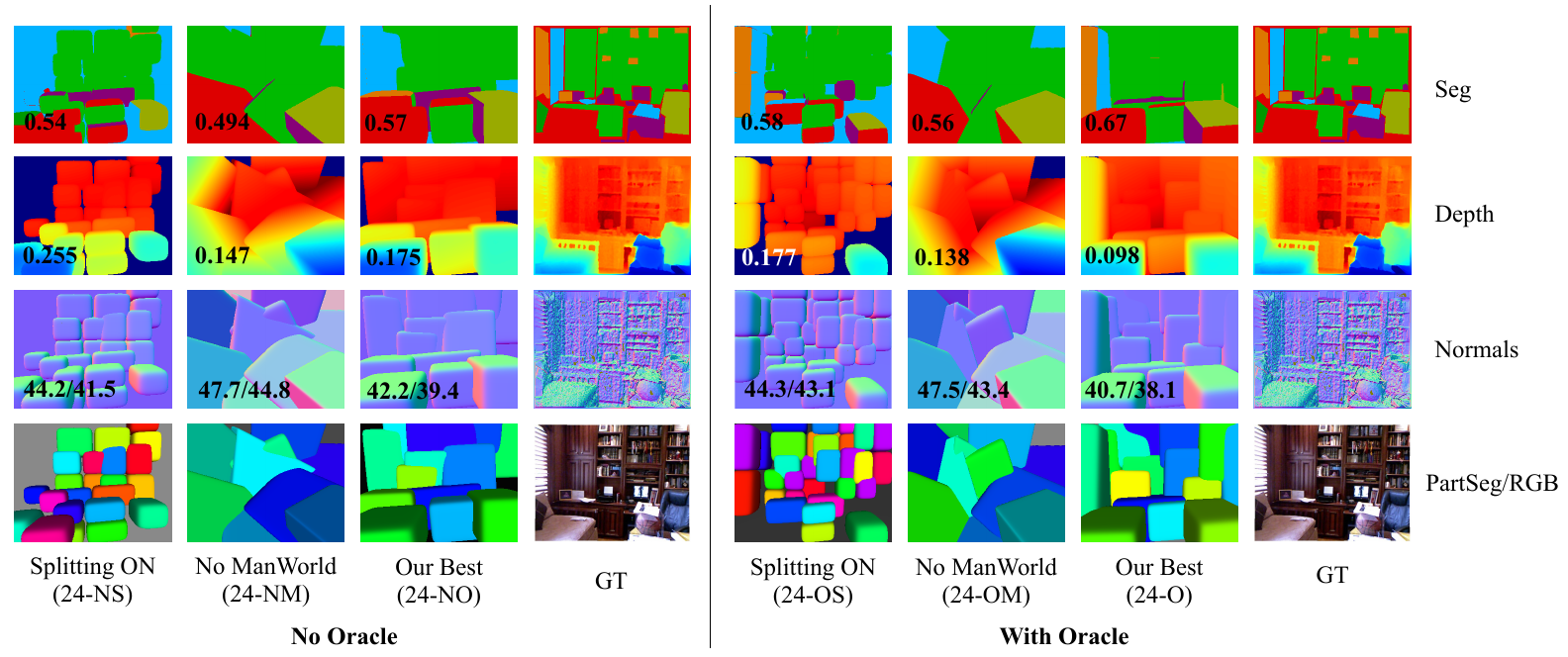} 
  
  \end{tabular}
  \caption{Qualitative ablation study on the effects of splitting during refinement and our Manhattan World losses. Results shown with and without ground truth depth and segmentation during refinement. Corresponding quantitative evaluation in Table \ref{tab:err1}. \textbf{Convex splitting} Applying convex splitting during the refinement step increases the granularity of our representation; however our optimization procedure struggles to improve the fit and convex pruning results in holes (comparing 24-NS against 24-NO and 24-OS against 24-O). \textbf{Manhattan World Losses} Removing the Manhattan World losses during training and refinement results in qualitatively cluttered representations. The depth is approximately correct, but the convexes are not organized in a manner representing the scene (e.g. maintaining parallel lines where possible). Our decompositions are quantitatively and qualitatively worse with the Manhattan World losses removed (comparing 24-NM against 24-NO and 24-OM against 24-O).} 
  \label{fig:spp1}
\end{figure*}

\begin{figure*}
  \centering
  \begin{tabular}{@{}c@{}}
    \includegraphics[width=.95\linewidth]{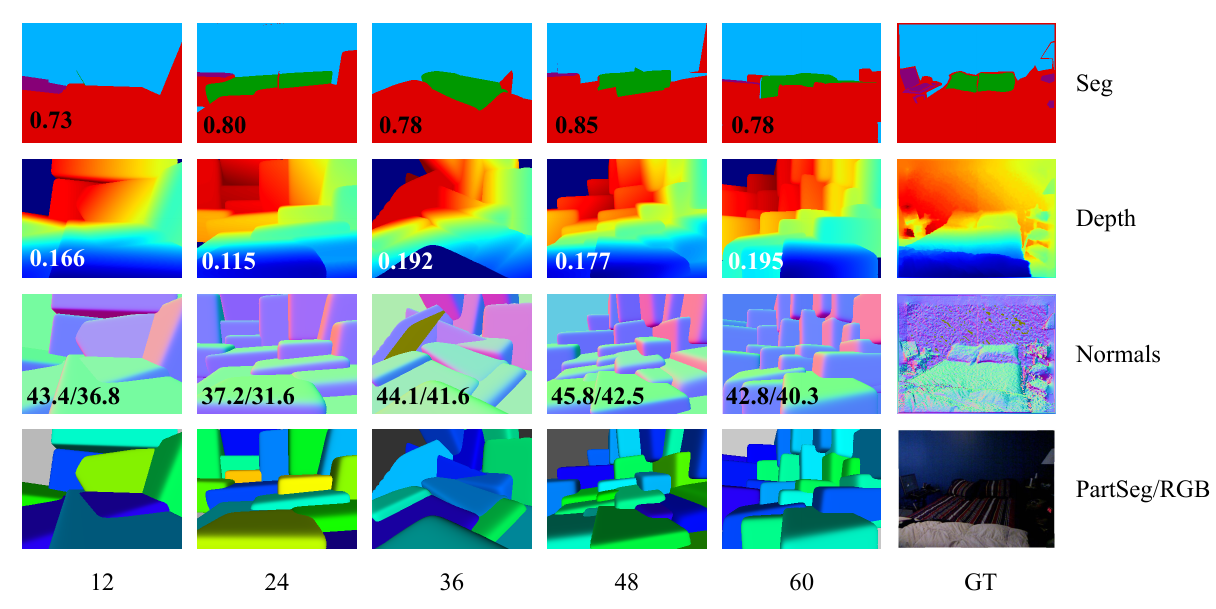}
  \end{tabular}
      \caption{Ablation study on number of parts at training time. We examine training with $n_{parts} \in [12,24,36,48,60]$ on a random NYUv2 test image. We refine without GT depth/seg. The optimal number of parts is this experiment was 24. Notice how less than this value, we run into bias issues, and above this value, we see variance problems. Quantitative results shown in Table \ref{tab:err1}.}
    \label{fig:nparts}
\end{figure*}

\begin{figure*}
  \centering
  \begin{tabular}{@{}c@{}}
    \includegraphics[width=.75\linewidth]{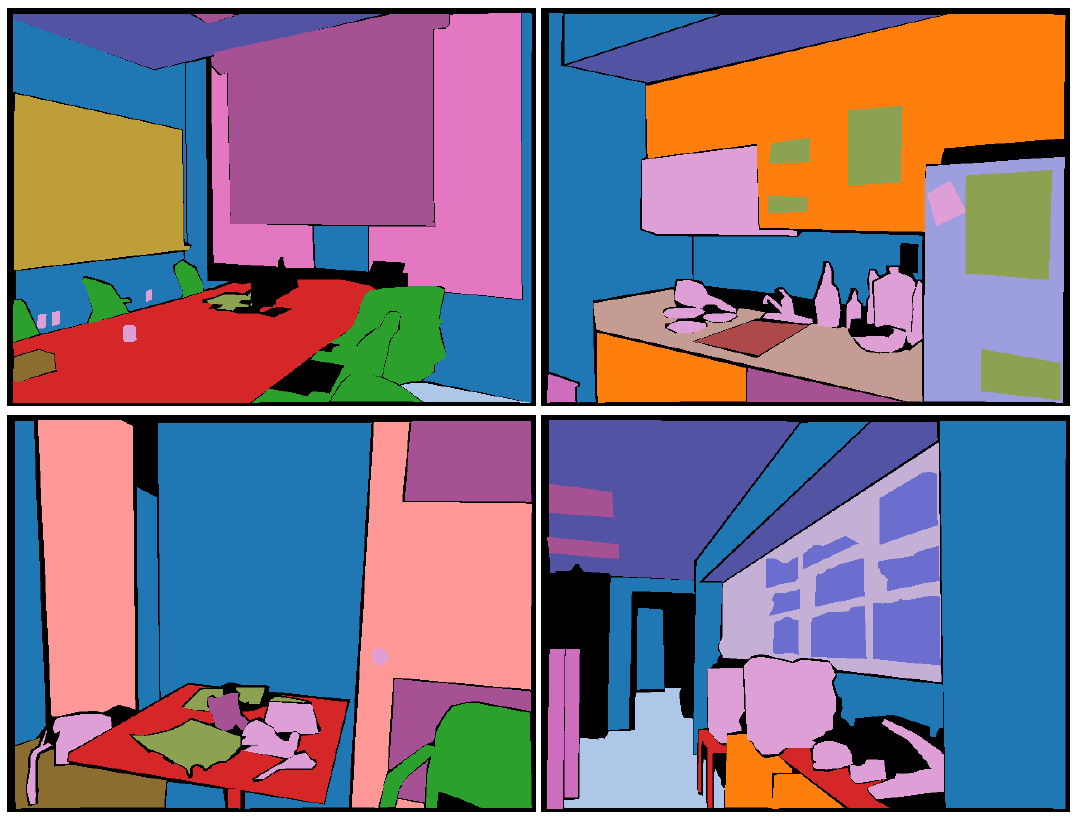}
  \end{tabular}
      \caption{Four example ground truth segmentation maps of NYUv2. The entropy loss encouraging our primitive decomposition to roughly obey segmentation boundaries quantitatively showed neutral results on this dataset - though it helped in very simple toy models we tried. Observe how the segmentations are quite cluttered due to occlusions or objects on top of one another. Say - posters on the fridge/cabinet or objects on a desk. The signal to the entropy loss can confuse the part decomposition in areas we want just one primitive representing that area. An obvious approach to dealing with scene complexity is more primitives. However, training with more primitives resulted in significant variance issues and a drop in quality beyond 24 as Table \ref{tab:err1} shows. Splitting the parts during refinement followed by polishing also failed to yield helpful results due to optimization difficulty. Finally, one could tackle this from the perspective of data: we could pre-process the segmentation with the intent to simplify them (similar to \cite{jiang2014finding}).}
    \label{fig:entropy}
\end{figure*}

In this supplement, we present additional ablations. We place quantitative evaluation in Table \ref{tab:err1}. First we examine the effects of adjusting the number of parts at training time, $n_{start} \in [12,24,36,48,60]$, finding that more parts does not necessarily translate to better quality due to bias-variance tradeoff issues (Fig. \ref{fig:nparts}). We also examine how our Manhattan World losses affect the quality of our primitives, finding that they help error metrics across the board. We show in Fig. \ref{fig:spp1} how the Manhattan World losses help orient the convexes in manner consistent with scene layout. Finally, we examine training and refining without our segmentation loss (entropy on the segmentation labels). Quantitatively, adding this entropy loss had an approximately neutral effect on our error metrics. We think that that this is due to the complexity of our segmentation labels and inability of our procedure to manage larger numbers of convexes (see Fig. \ref{fig:entropy}). 

\begin{figure*}
\centering
\includegraphics[height=3.6cm]{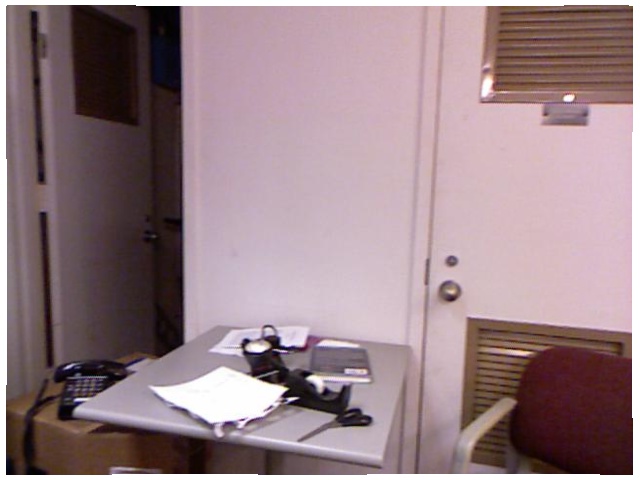}
\includegraphics[height=3.6cm]{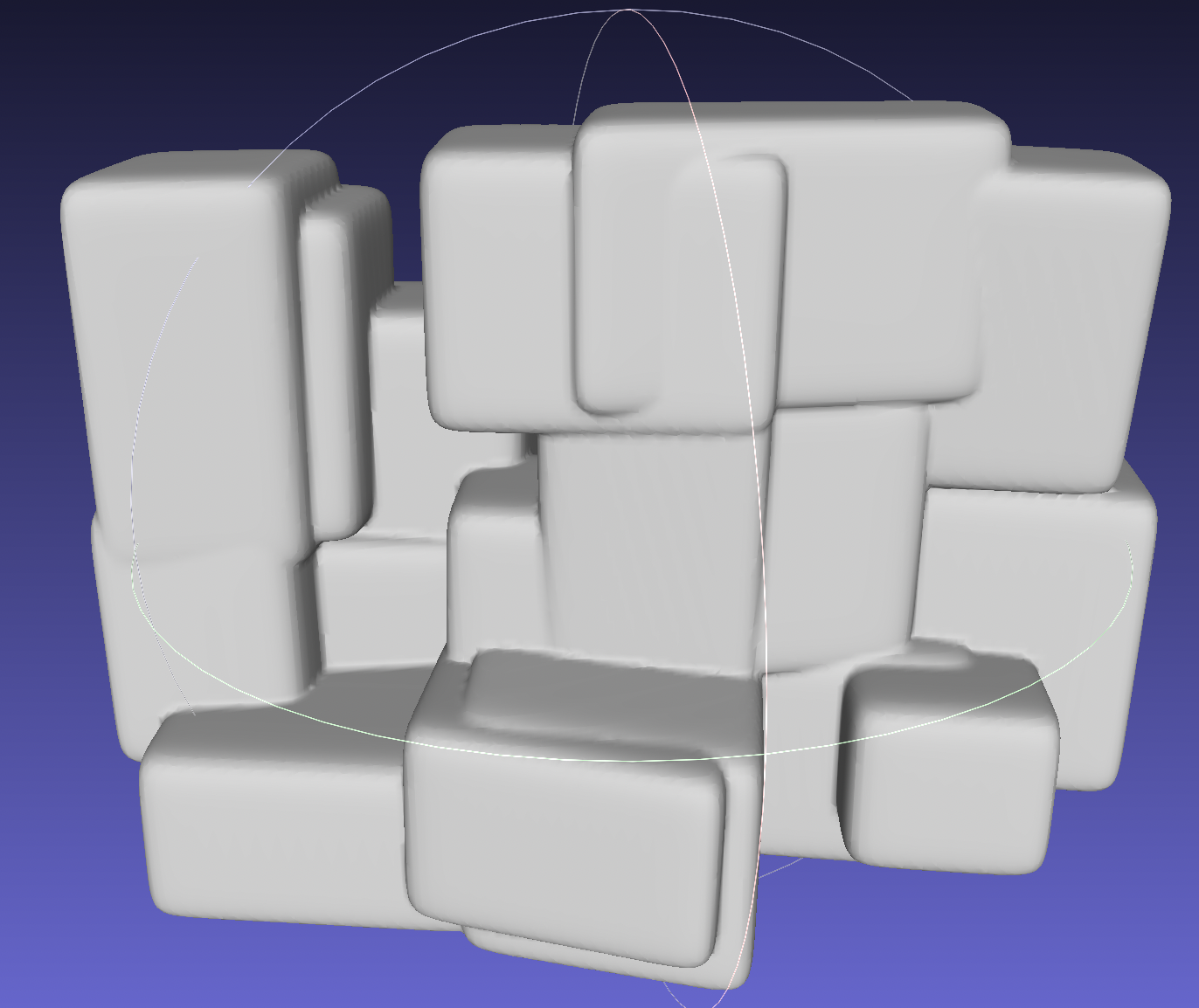}
\includegraphics[height=3.6cm]{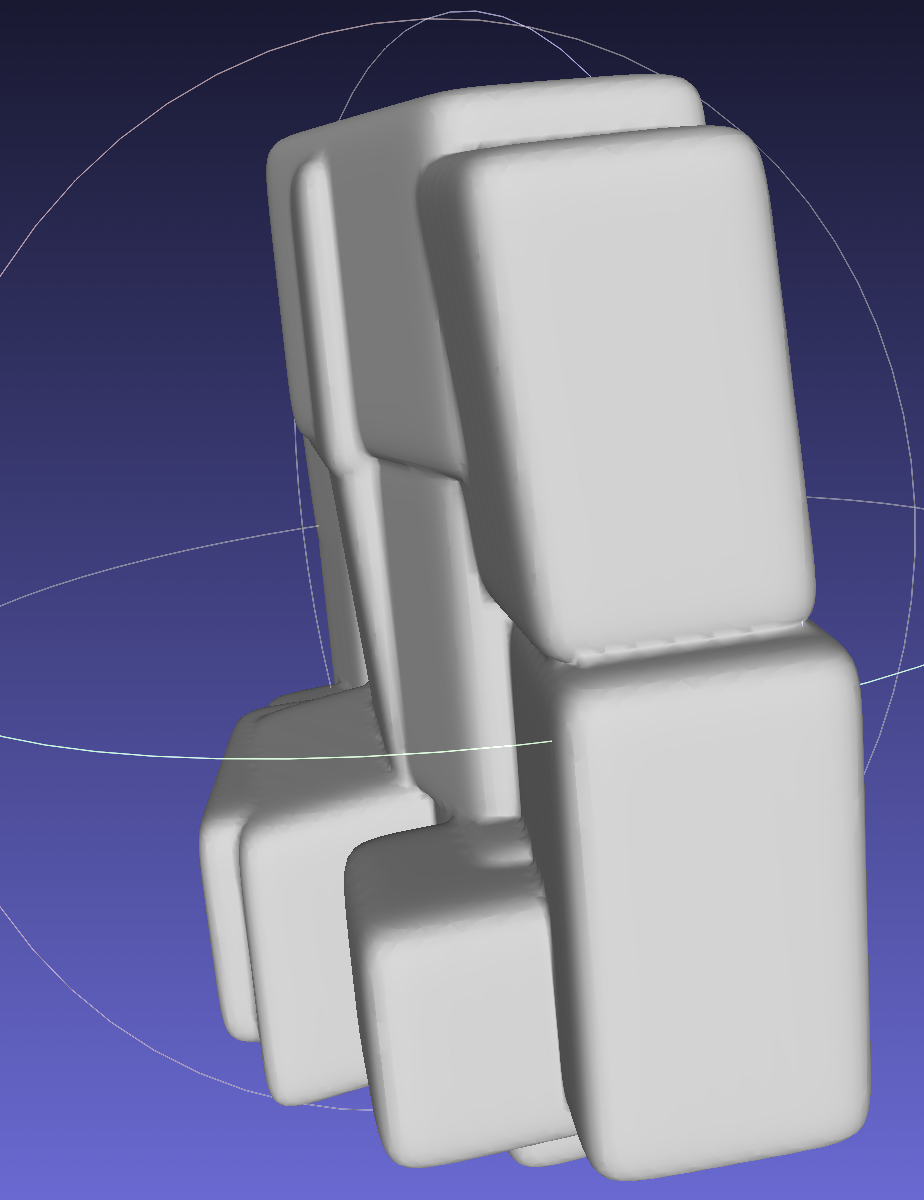}
\includegraphics[height=3.6cm]{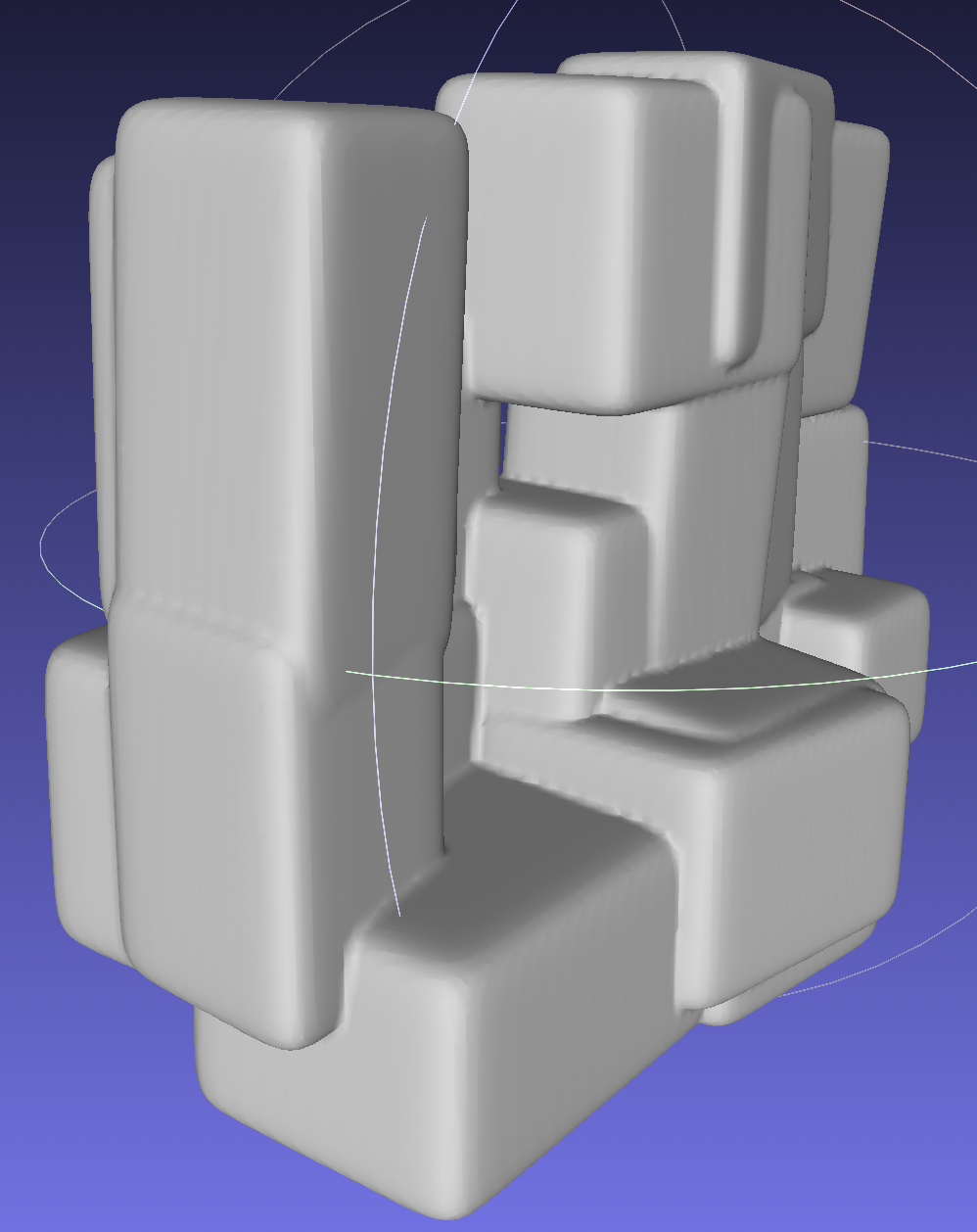}
\caption{Input image, front of primitives, right side, left side.}
\label{fig:view}
\end{figure*}

We show a 3D mesh of our primitives from multiple views in Fig.~\ref{fig:view}. We note that our primitives are represented in normalized coordinates and we preserve scale/shift coefficients in the X/Y/Z directions to raytrace our primitives from the original viewpoint and obtain a depth map in the original camera frame.

In Table~\ref{tab:compare}, we quantitatively compare our method against the most similar work, ~\cite{kluger2021cuboids}, using their error metric. Our AUC's are better across the authors' reported range $5-50cm$, but our mean is worse. That would indicate the presence of a few outlier test scenes that our method performed poorly on.

In Table~\ref{tab:err_abl}, we evaluate our method, removing one loss at a time. While the guidance loss had a marginal effect, the remaining losses meaningfully improved the error metrics (row C).

\vspace{4mm}

\begin{table*}
\resizebox{\textwidth}{!}{%
\begin{tabular}{c||c|c|c|c|c|c}
 & $AUC_{@50cm}$ & $AUC_{@20cm}$ & $AUC_{@10cm}$ & $AUC_{@5cm}$ & $mean_{cm}$ & $median_{cm}$\\
 \hline\hline
 \textbf{Ours - RGB} & \textbf{77.3} & \textbf{47.6} & \textbf{26.8} & \textbf{13.9} & $40.2$ & $26.2$\\
Kluger \textit{et al. } & $57.0$ & $33.1$ & $18.9$ & $10.0$ & \textbf{34.5} & -\\
\hline\hline
\textbf{Ours - Depth} & \textbf{86.9} & \textbf{72.5} & \textbf{56.5} & \textbf{38.2} & $26.6$ & $10.1$\\
Kluger \textit{et al.} & $77.2$ & $62.7$ & $49.1$ & $34.3$ & \textbf{20.8} & -\\
\hline\hline
\end{tabular}
}
\caption{Comparison with previous work~\cite{kluger2021cuboids} - Occlusion-Aware distance metric reported in all columns. Our AUC's are better, but mean is worse; medians indicate we suffer because of outliers.}
\label{tab:compare}
\end{table*}

\begin{table*}
\begin{center}
\resizebox{\linewidth}{!}{%
\begin{tabular}{||c | c c c c c c c|| c c || c c c c c || c ||} 
 \hline
  Cfg. & $n_{used}$ & AbsRel$\downarrow$ & RMSE$\downarrow$  & Mean$\downarrow$ & Median$\downarrow$ & $11.25^{\circ}\uparrow$ & $22.5^{\circ}\uparrow$ & $30^{\circ}\uparrow$ & $Seg_{Acc}\uparrow$ \\ 
 \hline\hline
no-unique & 16.7 & 0.157 & 0.695 & 38.092 & 34.501 & \textbf{0.153} & 0.328 & 0.437 & 0.613 \\
\hline
no-guidance & 14.4 & 0.145 & 0.624 & \textbf{37.422} & 33.471 & 0.149 & \textbf{0.339} & 0.453 & 0.611 \\
\hline
overlap-10 & 17.3 & 0.148 & 0.646 & 38.567 & 34.711 & 0.149 & 0.325 & 0.435 & \textbf{0.630} \\
\hline

no-overlap & 4.3 & 0.179 & 0.673 & 38.457 & 34.431 & 0.107 & 0.300 & 0.431 & 0.466 \\

\hline
no-volume & 16.2 & \textbf{0.144} & 0.618 & 37.462 & \textbf{33.292} & 0.141 & 0.338 & \textbf{0.454} & 0.623 \\
\hline
no-local & 1.0 & 0.499 & 1.856 & 87.673 & 82.785 & 0.006 & 0.019 & 0.034 & 0.212 \\
\hline\hline
C & 14.4 & \textbf{0.144} & \textbf{0.603} & 38.235 & 33.621 & 0.133 & 0.335 & 0.451 & 0.615 \\
 \hline\hline
\end{tabular}}

\caption {Ablations with one CVXnet loss removed at a time. All are non-oracle. C is our final non-oracle model (Table~\ref{table:err}). Depth metrics worse without unique param. loss (eqn. 5 of CVXnet). Guidance loss (eqn. 6) has an approx. neutral effect. Removing the overlap loss (decomp. loss eqn. 4 in CVXnet) harms quality; scaling it up to $10$, beyond the default $0.1$, slightly hurts error metrics, likely because convexes need to move freely during the training process. It remains future work to completely eliminate overlaps while preserving quality. Removing the volume loss has an approx. neutral effect on the error metrics though parsimony is harmed. The localization loss (eqn. 7) is critical to the method.
}
\label{tab:err_abl}
\end{center}

\end{table*}

\end{document}